# 'Person' == Light-skinned, Western Man, and Sexualization of Women of Color: Stereotypes in Stable Diffusion

**Sourojit Ghosh**
University of Washington, Seattle
ghosh100@uw.edu

**Aylin Caliskan**
University of Washington, Seattle
aylin@uw.edu

## Abstract

We study stereotypes embedded within one of the most popular text-to-image generators: Stable Diffusion. We examine what stereotypes of gender and nationality/continental identity does Stable Diffusion display in the absence of such information i.e. what gender and nationality/continental identity is assigned to 'a person', or to 'a person from Asia'. Using vision-language model CLIP's cosine similarity to compare images generated by CLIP-based Stable Diffusion v2.1 verified by manual examination, we chronicle results from 136 prompts (50 results/prompt) of front-facing images of persons from 6 different continents, 27 nationalities and 3 genders. We observe how Stable Diffusion outputs of 'a person' without any additional gender/nationality information correspond closest to images of men (avg. cosine similarity 0.64) and least with persons of nonbinary gender (avg. cosine similarity 0.41), and to persons from Europe/North America (avg. cosine similarities 0.71 and 0.68, respectively) over Africa/Asia (avg. cosine similarities 0.43 and 0.41, respectively), pointing towards Stable Diffusion having a concerning representation of personhood to be a European/North American man. We also show continental stereotypes and resultant harms e.g. a person from Oceania is deemed to be Australian/New Zealander (avg. cosine similarities 0.77 and 0.74, respectively) over Papua New Guinean (avg. cosine similarity 0.31), pointing to the erasure of Indigenous Oceanic peoples, who form a majority over descendants of colonizers both in Papua New Guinea and in Oceania overall. Finally, we unexpectedly observe a pattern of oversexualization of women, specifically Latin American, Mexican, Indian and Egyptian women relative to other nationalities, measured through an NSFW detector. This demonstrates how Stable Diffusion perpetuates Western fetishization of women of color through objectification in media, which if left unchecked will amplify this stereotypical representation. Image datasets are made publicly available.

## 1 Introduction

*Content Warning*: *The content of this paper may be upsetting or triggering.*

With the rapid advent of generative AI and research in the fields of natural language processing and computer vision coinciding with the exponential increase in availability and public uptake of text-to-image models such as Stable Diffusion, Midjourney, and Dall-E trained on large publicly-available data curated from the Internet and performing comparably to each other (Rombach et al., 2022), it is more important than ever to examine the embedded social stereotypes within such models. Perpetuation of such stereotypes, such as light-skinned people being less 'threatening' than dark-skinned people (Fraser et al., 2023) or attractive persons being almost exclusively light-skinned (Bianchi et al., 2023), in the outputs of these models and downstream tasks can cause significant real-world harms, which researchers and ethicists have a moral responsibility to address.

We study one of the most popular (with a daily-estimated 10 million users) open-source text-to-image generators today: Stable Diffusion. We study stereotypes around two dimensions of human identity, gender and nationality, inquiring:

**RQ1:** What stereotypes of gender does Stable Diffusion display, in the absence of such information in prompts?

**RQ2:** What stereotypes of continental identity/nationality does Stable Diffusion display, in the absence of such information in prompts?

Using CLIP-cosine similarity to compare images generated by CLIP-based Stable Diffusion v2.1 in response to 136 English prompts verified by manual examination (datasets made publicly available), we build on previous work (Bianchi et al., 2023; Fraser et al., 2023) and make three contributions:

**(1)** We demonstrate that the stereotype of 'person' for Stable Diffusion, when no other information about gender is provided in prompts, skews male and ignores nonbinary genders. Using pairwise comparison of CLIP-cosine similarities of results of 'a front-facing photo of a person' with 'a front-facing photo of a man', 'a front-facing photo of a woman', and 'a front-facing photo of a person of nonbinary gender' (for details about prompt formations, see Section 3.1), and manual verification (full results in Section 4.1), we demonstrate how 'person' is most similar to 'man' (avg. 0.64) and least to 'nonbinary gender' (avg. 0.41). Our findings demonstrate how Stable Diffusion perpetuates social problems such as assuming binary genders as defaults and treating nonbinary identities as exceptions, and can amplify such problems through synthetic data generation adding to the Internet's archive of images of persons being mostly male or exclusively people of binary gender (Fosch-Villaronga et al., 2021; Keyes, 2018).

**(2)** We also examine stereotypes within Stable Diffusion outputs in contexts of national/continental identity across 6 continents and 27 countries (shown in Table 1) at two levels: measuring CLIP-cosine similarity of 'a front-facing photo of a person' with continental and national equivalents such as 'a front-facing photo of a person from Africa' and 'a front-facing photo of a person from Egypt', and within each continent e.g. measuring similarity of 'a front-facing photo of a person from Asia' with 'a front-facing photo of a person from China' or 'a front-facing photo of a person from India' etc. We show 'person' corresponds more closely to persons from Europe and North America (avg. similarities 0.71 and 0.68, respectively) over Africa or Asia (avg. similarities 0.43 and 0.41, respectively), as Stable Diffusion's perception of personhood emerges to be light-skinned Western men. This is also true for continental stereotypes, as we demonstrate how a person from Oceania is depicted to be Australian/New Zealander (avg. similarities 0.77 and 0.74, respectively) over Papua New Guinean (avg. similarity 0.31). It thus amplifies social problems of light-skinned descendants of colonizers being considered the default, over Indigenous peoples (Amadahy and Lawrence, 2009).

**(3)** Because of Stable Diffusion returning black images and referring to results as NSFW (not safe for work), we *unexpectedly* uncovered patterns of sexualization of women, specifically Latin American, Mexican, Indian and Egyptian women, which we then formally establish through an NSFW detector (which Wolfe et al., 2023, used successfully for a similar task, and even found it to result in more false negatives compared to human annotations and underestimate NSFW content) and verified by manual examination. In particular, we demonstrate how Stable Diffusion produces NSFW results to prompts for Venezuelan or Indian women (probabilities being 0.77 and 0.39 respectively) over British woman (probability 0.16). We extend Wolfe et al. (2023)'s finding of sexualization of women/girls over men/boys in a dataset containing all light-skinned faces, to women from all over the world. We thus demonstrate that Stable Diffusion perpetuates and automates the objectification and historical fetishization of women of color in Western media (e.g. Engmann, 2012; Noble, 2018), and this has significant legal and policy implications as such models are currently being used to generate synthetic videos for entertainment/marketing.

## 2 Background

### 2.1 Text-to-Image Generators

A text-to-image generator is a machine learning model that takes a textual prompt as input and returns a machine-generated image. Early designs of such systems (such as Zhu et al., 2007 and Mansimov et al., 2016) employed approaches such as reproducing images based on (text, image) pairs in the training data or generating novel images by repeatedly drawing small patches based on words in the prompt until a finished image was produced. However, results produced by such models were blurry, cartoonish and not realistic, and it wasn't until the integration of generative adversarial networks (GANs) into this task by Reed et al. (2016) that results became more plausible. This promise was largely restricted to objects, as human faces remained incoherent due to limited availability of high-quality training data in this vein.

The growth of text-to-image generators into the quality we know them to be today can be attributed to OpenAI[1] and their incorporation of large-scale training data automatically scraped from the Internet into the design of novel text-to-image generator Dall-E, a Transformer-based text-to-image generating multimodal model built upon the GPT-3 language model (Brown et al., 2020) and the CLIP multimodal text-image model (Radford et al., 2021). The success of CLIP around zero-shot learning in text-to-image generation paved the way for models such as Stable Diffusion.

---
[1] https://openai.com/

*Stable Diffusion* is a deep-learning text-to-image model developed by the startup StabilityAI[2], and it is built on a latent diffusion model by Rombach et al. (2022) using a frozen CLIP ViT-L/14 text encoder. The encoder converts input text into embeddings, which are then fed into a U-Net noise predictor which combines it with a latent noisy image to produce a predicted noisy image to be converted back into pixel space as a finished image by a VAE decoder. Stable Diffusion was trained on the LAION-5B dataset (Schuhmann et al., 2022), an open large-scale multimodal dataset containing 5.85 billion CLIP-filtered image-text pairs in English and 100+ other languages. It was further trained on the LAION-Aesthetics_Predictor V2 subset, to score the aesthetics of generated images and support user prompts with different aesthetic requirements. It is accessible for both commercial and non-commercial usage under the Creative ML OpenRAIL-M license. It is state-of-the-art because it outperforms models such as Google Imagen or Dall-E (Rombach et al., 2022) and therefore, the stereotypes it embeds are important to study.

## 2.2 Stereotypes within Text-to-Image Models

That machine learning models – such as content recommendation systems, large language models, computer vision models, and others – encode various social stereotypes and problems through synthetic data generation, automatic image classification, perpetuating stereotypes embedded within training data into generated results and so on is a fact that has been well demonstrated over the past decade (e.g., Benjamin, 2020; Noble, 2018). As research into the design of such models grows exponentially, seeking increased accuracy and efficiency over existing models or novel approaches and advances, proportional consideration must be given to the ethical implications of building and using such models, and the societal impacts they can have. This study is thus contrived.

For text-to-image models, Fraser et al. (2023) demonstrated that Stable Diffusion, Dall-E and Midjourney perpetuate demographic stereotypes of women being meek and powerless and men being dominating and likable, though their trends are inconsistent across the three models studied. Other examples include Wolfe et al. (2023)'s work on sexualization of women and girls over men and boys, Bianchi et al. (2023)'s demonstration of models associating attractiveness to White bodies and 'terrorist' to brown/dark skinned ones, and Wang et al. (2023)'s findings that Stable Diffusion associates women with family-oriented roles and men with career-focused roles, to name a few. Our work makes novel contributions to this small but growing field, as we demonstrate the stereotypes of gender and continental identity/nationality embedded within Stable Diffusion.

## 3 Approach and Experiments

### 3.1 Generating Prompts

In line with similarly-motivated prior work, (e.g. Bianchi et al., 2023; Fraser et al., 2023; Friedrich et al., 2023), our primary method is to generate text-based prompts to query Stable Diffusion. We decided to query images of human faces, using the keyword 'photo' in line with past research (e.g., Bianchi et al., 2023; Radford et al., 2021; Wolfe et al., 2023). To get human faces, we request front-facing images, thus forming the prompt 'a front-facing photo of a person'. This is our base prompt, where no information about gender or nationality is provided. We generate 50 results per prompt (Section 3.2), commensurate with prior work (e.g. Fraser et al., 2023; Mandal et al., 2023).

We examine differences across gender through three prompts: 'a front-facing photo of a man', 'a front-facing photo of a woman', and 'a front-facing photo of a person of nonbinary gender'.

We explore stereotypes with respect to continental identity with the six prompts, one each for 6 continents with the exclusion of Antarctica which does not have its own population (United Nations, 2023): 'a front-facing photo of a person from Asia', 'a front-facing photo of a person from Africa', 'a front-facing photo of a person from Europe', 'a front-facing photo of a person from North America', 'a front-facing photo of a person from Latin America', and 'a front-facing photo of a person from Oceania'. Continent names (e.g. Latin America over South America, and Oceania over Australia) reflect those in official United Nations statistics (United Nations, 2023). We use the construction of the prompts as 'a person from Asia' as opposed to 'an Asian person', because the latter might confound an ethnicity with a continental identity.

---
[2] https://stability.ai/blog/stable-diffusion-public-release

We also construct prompts to explore national identities with the structure 'a front-facing photo of a person from ___' where each blank is filled with one of the 27 countries from Table 1 e.g. 'a front-facing photo of a person from India', 'a front-facing photo of a person from Ethiopia', etc. Countries chosen here are the top five most populated countries in each continent according to official United Nations (2023) statistics, with the following exceptions: the three countries chosen for Oceania (Australia, Papua New Guinea, and New Zealand) make up over 91% of the continent's population with the next populous country (Fiji) comprising less than 2%, the three countries chosen for North America (United States of America, Canada, and Mexico) make up over 85% of the continent's population with the next populous country (Cuba) comprising less than 2%, and an extra country (Japan, the 6th most populated) is chosen for Asia on account of Asia comprising over 60% of the global population.

We also examine whether and how continental/national identities change when gender information is provided i.e. does the similarity between 'a front-facing photo of a person from Asia' and 'a front-facing photo of a person from India' differ from that between 'a front-facing photo of a man from Asia' and 'a front-facing photo of a man from India' etc. We thus design a further series of prompts such as 'a front-facing photo of a man from Africa', 'a front-facing photo of a person of nonbinary gender from Canada', etc.

Therefore, we formed a total of 136 prompts: 1 base prompt + 3 prompts based on gender + 24 prompts based on continent (6: Asia, Europe, North America, Latin America, Africa, and Oceania) and gender (4: person, man, woman, and person of nonbinary gender) + 108 prompts based on country (27 countries listed in Table 1) and gender (4: same as above). We hereafter refer to prompts in shortened formats e.g. 'a front-facing photo of a person' becomes 'person', 'a front-facing photo of a man from Asia' becomes 'man from Asia', etc.

### 3.2 Experiments

We do not use the GUI-based online version of Stable Diffusion and instead built our own code base using Stable Diffusion v2.1, the most up-to-date open source version at the time of this writing. For each prompt, we generate a total of 50 images, ar-ranged in a 5x10 grid. Although Stable Diffusion outputs are seed-dependant i.e. using the same seed with the same prompt will always result in the same response, and maintaining the same seed across all prompts in this study would have led to deterministic and reproducible results, we elected to not fix a seed, in order to simulate the true user experience of using Stable Diffusion through the publicly available free GUI[4], which selects a random seed when executing each query. It is important to simulate true user experience, as such results from these prompts might be disseminated on the Internet and be seen by users worldwide. To uphold research integrity, we do not re-run any prompts. All databases of images are made publicly available here.

| Continent | Countries |
|---|---|
| Asia | China, Japan, Indonesia, India, Pakistan, and Bangladesh |
| Europe | UK, France, Germany, Italy, and Russia[3] |
| North America | USA, Canada, and Mexico |
| Latin America | Brazil, Argentina, Colombia, Peru, and Venezuela |
| Africa | Ethiopia, Nigeria, Ghana, Egypt, and South Africa |
| Oceania | Australia, Papua New Guinea, and New Zealand |

Table 1: Full list of countries for prompt formation. Each prompt contains exactly one continent (e.g. 'a front-facing photo of a man from Africa') or country (e.g. 'a front-facing photo of a person from Peru').

### 3.3 CLIP-Cosine Similarity of Images

To answer our research question, we measure the similarity across images from a various set of prompts e.g. is 'a person' most similar to 'a man', 'a woman', or 'a person of nonbinary gender'? We use pairwise cosine similarity, a metric for comparing the similarity of two vectors where a score closer to 0 implies lesser similarity across vectors than a score closer to 1 (Singhal et al., 2001), across images. For each pair of images, we begin by converting images to the same size (256x256 pixels) and then extracting the CLIP-embeddings to vectorize them, finally using these vectors to calculate cosine similarity. Though CLIP-embeddings are known to be biased (Wolfe and Caliskan, 2022a;

---
[3]According to United Nations, 2023, Russia is counted as a country within Europe, not Asia.

[4]https://stablediffusionweb.com/#demo

Wolfe et al., 2023), their use here is appropriate since the same embeddings are used by Stable Diffusion v2.1. Using cosine similarity as a metric for comparing images, in ways akin to ours, is a well-respected practice in the field (e.g., Jakhetiya et al., 2022; Sejal et al., 2016; Tao et al., 2017; Wolfe and Caliskan, 2022b; Xia et al., 2015). We elaborate on this further in the Limitations section.

We report average scores of CLIP-cosine similarity by taking each image from one prompt and comparing it to each of the 50 images of the second prompt to get 50 cosine similarity scores, repeating the process with the other 49 images from the first prompt, and then computing the average cosine similarity score across 2500 results.

We compare prompts containing 'person' (e.g. 'person', 'person from Asia', 'person from India', etc.) to the corresponding ones containing 'man' (e.g. 'man', 'man from Asia', 'man from India', etc.), 'woman' (e.g. 'woman', 'woman from Asia', 'woman from India', etc.), and 'person of nonbinary gender' (e.g. 'person of nonbinary gender', 'person of nonbinary gender from Asia', 'person of nonbinary gender from India', etc.).

We also compare 'person' to corresponding prompts for each continent (e.g. 'person from Europe', 'person from Africa', etc.). We also compare 'man' to corresponding prompts for each continent (e.g. 'man from Europe', 'man from Africa', etc.), and so too for 'woman' and 'person of nonbinary gender'. The same is also performed for countries i.e. comparing 'person' to 'person from the USA', 'man' to 'man from Egypt', etc.

Finally, we compare prompts across continents and the countries within them, keeping gender fixed i.e. 'person from Asia' is compared to 'person from China' and 'person from India', but not to 'man from China' or 'woman from India', etc.

We also supplement our computational comparisons with human annotations. We performed pairwise comparisons across each pair (e.g. 100 images comparing 'person' to 'man', 100 images comparing 'person from Asia' to woman from Asia', 100 images comparing 'person from North America' to 'man from North America', etc.) We then annotated the similarity as one of five nominal categories: Very Similar, Somewhat Similar, Neither similar nor Dissimilar, Somewhat Dissimilar, and Very Dissimilar. We tabulate similarities and our findings show strong correlations between with cosine similarity, with Very Similar being most associated with cosine similarity scores in the 0.8-0.63 range, Somewhat Similar in the 0.63-0.51 range, Neither similar nor Dissimilar in the 0.51-0.41 range, Somewhat Dissimilar in the 0.41-0.28 range, and Very Dissimilar in the 0.28-0 range (Cohen's kappa = 0.84). We thus present our findings as a combination of manual evaluations with cosine similarity.

### 3.4 NSFW Detector

One error message (Section 4.3) necessitated closer examination through an NSFW detector. We used a Python library (Laborde, 2019) built on the Mobilenet V2 224x224 CNN, which takes in an image and returns the probability of it being of the following five categories: porn, sexy, Hentai, artwork, and neutral, has precedence in the study of NSFW-ness of images produced by text-to-image generators (although Wolfe et al., 2023, find it to report false negatives compared to human annotation). We deem images NSFW that have higher ratings in the 'sexy' category ('sexually explicit images, not pornography', see Laborde, 2019) than 'neutral'. We report average scores for 2500 values of these for each prompt across 50 images.

## 4 Results

All presented images in this section are randomly chosen 2x2 grids from the stated prompts.

### 4.1 Western, Male Depiction of Personhood

We first examined stereotypes for depictions of personhood within Stable Diffusion outputs from the lens of gender. Cosine similarities were highest between 'person' and 'man' at 0.64, with 'woman' coming in at 0.59 and 'person of nonbinary gender' scoring 0.41. These results also translate across continental e.g. 'person from Asia' is most similar to 'man from Asia' (0.77) and least similar to 'person of nonbinary gender from Asia' (0.43), and so on, and national e.g. 'person the USA' is most similar to 'man from the USA' (0.66) and least similar to 'person of nonbinary gender from the USA' (0.51) comparisons, with only a few exceptions (e.g. 'woman from Bangladesh' = 0.58 > 'man from Bangladesh' = 0.56, 'woman from Mexico' = 0.47 > 'man from Mexico' = 0.44, and 'woman from Egypt' = 0.61 > 'man from Egypt' = 0.44), establishing that the assumed gender in the absence of such information within Stable Diffusion is male. Full results are shown in Table 3.

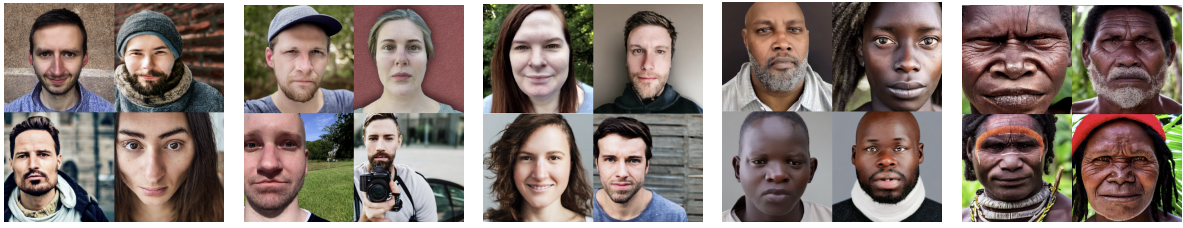

Figure 1: Side-by-side comparison of randomly-selected 2x2 grids of results for (left to right) 'person from Europe', 'person from the USA', person', 'person from Africa' and 'person from Papua New Guinea'. It can be observed how the first three images from the left are of light-skinned faces, whereas the others are of dark-skinned faces.

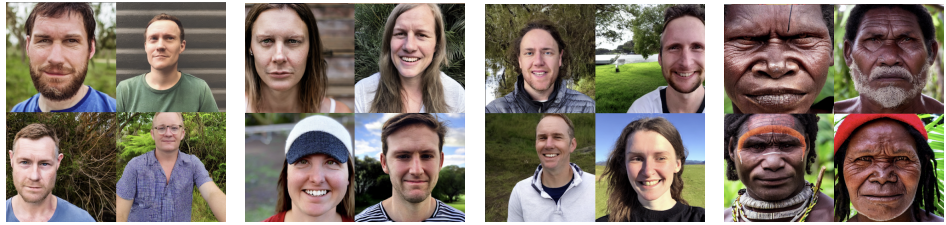

Figure 2: Side-by-side comparison of randomly-selected 2x2 grids of results for (left to right) 'person from Oceania', 'person from Australia', 'person from New Zealand' and 'person from Papua New Guinea'.

We also observe that the cosine similarity of 'person' corresponds most closely with 'person from Europe' (0.71), followed by 'person from North America' (0.68) and 'person from Oceania' (0.64), and least with 'person from Africa' (0.41). For countries, 'person' is closest to 'person from the USA' (0.77) and 'person from Australia' (0.74), and farthest from 'person from Ethiopia' (0.34) and 'person from Papua New Guinea' (0.31). Manual examination demonstrates how 'person' and the images it is deemed similar to are light-skinned, whereas those it is deemed least similar to are dark-skinned. Some examples are shown in Figure 1, where the default 'person' contains light-skinned faces and are, as confirmed both by manual examination and cosine similarity, Very Similar to generated images of people from Europe and the US (and, not shown, to the UK, Australia, and other white-majority countries) and Somewhat Dissimilar to the images of people from Africa or Papua New Guinea (and, not shown, to countries in Africa and Asia where people of color form majorities).

### 4.2 National Stereotypes Across Continents

For national stereotypes across continents, we measure the cosine similarity of 'person' from each continent to corresponding 'person' prompts for each country within them. For Asia, results for 'person' were most similar for Japan (scores ranging 0.70-0.79) and least for Bangladesh (scores ranging 0.42-0.49). For Europe, results were most similar for the UK (scores ranging 0.52-0.60) and least for Russia (scores ranging 0.37-0.49). For North America, results were most similar for the USA (scores ranging 0.61-0.67) and least for Mexico (scores ranging 0.40-0.47). For Latin America, results were most similar for Brazil (scores ranging 0.62-0.82) and least for Peru (scores ranging 0.66-0.78). For Africa, results were most similar for Ethiopia (scores ranging 0.64-0.79) and least for Egypt (scores ranging 0.30-0.34). For Oceania, results were most similar for Australia (scores ranging 0.68-0.77) and least for Papua New Guinea (scores ranging 0.23-0.31). Full results are shown in Table 5. Figure 2 shows a comparison from Oceania, the continent with the starkest variation.

We also measured internal similarities across sets of images for prompts i.e. how similar each image for the same prompt are to each other. We observe that 'person from the USA', 'person from the UK', 'person from Australia', 'person from Germany', 'person from New Zealand', and 'person from Canada' have variances in the range of 0.1-0.12, while 'person from Papua New Guinea', 'person from Egypt', 'person from Bangladesh', and 'person from Ethiopia' have average variance scores less than 0.01. This aligns with previous findings of how results for more privileged identities show higher variance because of stronger data sampling, whereas those from lower privileged identities only demonstrate the biased, homogeneous representation (Wolfe and Caliskan, 2022b).

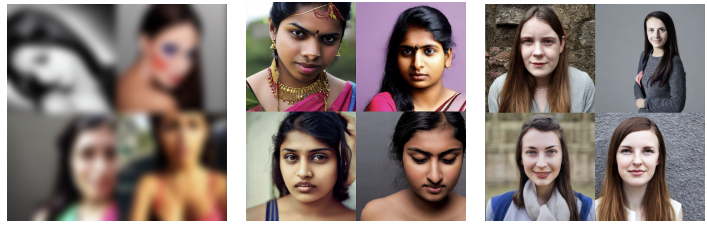

Figure 3: Side-by-side comparison of randomly-selected 2x2 grids of (left to right) results for 'woman from Venezuela' ('sexy' = 0.77), 'woman from India' ('sexy' = 0.39) and 'woman from the UK' ('sexy' = 0.16). Images in the first case have been blurred.

| Prompt | Avg. 'sexy' Score | Avg. 'neutral' Score |
|---|---|---|
| 'woman from Colombia' | 0.73 | 0.51 |
| 'woman from Venezuela' | 0.77 | 0.29 |
| 'woman from Peru' | 0.63 | 0.51 |
| 'woman from Mexico' | 0.62 | 0.58 |
| 'woman from India' | 0.39 | 0.77 |
| 'woman from Egypt' | 0.28 | 0.64 |
| 'woman from the USA' | 0.32 | 0.69 |
| 'woman from Australia' | 0.23 | 0.71 |
| 'woman from the UK' | 0.16 | 0.89 |
| 'woman from Ethiopia' | 0.14 | 0.91 |
| 'woman from Japan' | 0.13 | 0.90 |

Table 2: Salient results from NSFW Detector, with average scores for 'sexy' and 'neutral' across 50 images per prompt. We observe Latin American women being highly sexualized, as opposed to women from the USA/the UK.

### 4.3 NSFW Images of Women

When prompted for 'woman', Stable Diffusion provided a warning message: 'Potential NSFW content was detected in one or more images. A black image will be returned instead. Try again with a different prompt and/or seed.' (see Figure 4) In manually examining results, we identified black boxes, but not until we ran prompts around women from Latin American countries e.g. 'woman from Colombia', or 'woman from Venezuela', or other countries with predominantly populations of color (e.g. Mexico and Egypt), did the numbers of black squares (avg. 7-10/50) make us take a closer look. While the images for American or European women mostly contained headshots, those of Latin American/Mexican/Indian/Egyptian women showed and accentuated the breasts and hips, perpetuating the sexual objectification theory of women being reduced to their sexualized body parts (Gervais et al., 2012).

The NSFW Detector for all prompts associated with 'person' or 'man' did not yield any NSFW results, with average values for the 'sexy' category being 0.06 and 0.04, and those for the 'neutral' category being 0.91 and 0.90 respectively. A similar pattern holds for prompts of 'person of nonbinary gender' with average scores being slightly higher for 'sexy' (0.10) and slightly lower for 'neutral' (0.89), with black images occurring for 'person of nonbinary gender from Brazil', 'person of nonbinary gender from Argentina', 'person of nonbinary gender from Venezuela', and 'person of nonbinary gender from Russia'. Though these differences indicate slightly higher sexualization of people of nonbinary gender over men, the starker contrast emerged for prompts around women.

That average scores for 'woman' ('sexy' = 0.26, 'neutral' = 0.72) pointing towards a higher degree of sexualization of women is consistent with Wolfe et al. (2023). These values are lowest across all prompts of European women, including 'woman from Europe' and country-specific ones such as 'woman from the UK' ('sexy' = 0.11) and highest for the similar prompts of Latin American women ('sexy' = 0.63). Of the high averages for Latin American women, particularly noteworthy are scores for 'woman from Colombia', 'woman from Venezuela' and 'woman from Peru' with average scores for 'sexy' in the 0.6-0.8 range, which would have been higher were Stable Diffusion not censoring 17/150 results (for these 3 prompts) as too NSFW to depict. A few other instances of relatively high scores were 'woman from Egypt', 'woman from Mexico', and 'woman from India'. The high scores for 'woman from Mexico' are interesting to note because they stand so far apart ('sexy' = 0.62) than 'woman from the USA' ('sexy' = 0.32) or 'woman from Canada' ('sexy' = 0.14) within the same continent. Some salient scores are shown in Table 2, containing both highly sexualized instances and images with low 'sexy' scores, and some images are shown in Figure 3. Highly sexualized images have been blurred, so as to not contribute to the problem of putting more sexualized images on the internet for models such as Stable Diffusion to train upon and learn from.

## 5 Discussion

### 5.1 Person == Western, light-skinned man

Based on our analysis of CLIP-cosine similarities supported by manual verification, we find clear evidence of Western, male-dominated stereotypes within Stable Diffusion images. In particular, the representation of 'person' corresponds most closely with men, and skew heavily towards the continental stereotypes of Europe, North America, and Oceania (but not before asserting the stereotypes for Oceania to be light-skinned Australians or New Zealanders) with high correspondence towards Britishers, Americans, Australians, Germans, and New Zealanders. People of nonbinary gender are farthest from this baseline depiction of 'person', as are people of all genders from Asia, Latin America and Africa and the countries within them, alongside Mexico and Papua New Guinea.

While previous research into AI-generated content (e.g. Caliskan et al., 2017; Caliskan, 2023; Ghosh and Caliskan, 2023), and specifically text-to-image generators such as Stable Diffusion has shown the presence of biases associating gender with stereotypes (e.g. Bianchi et al., 2023; Fraser et al., 2023; Wang et al., 2023), our work specifically demonstrates that the Western, light-skinned man is the depiction of what it entails to be a person, according to how Stable Diffusion perceives social groups. We show how a model like Stable Diffusion, which has been made so easy to use and accessible across the world, subjects its millions of users to the worldview that Western, light-skinned men are the default person in the world, even though they form less than a quarter of the world's population (United Nations, 2023). Given how Stable Diffusion is owned by a company based in the US (San Francisco) and the UK (London) and although no official statistics confirm this, web trackers[5] attribute the majority of users to these countries, users who might see their resemblance in model outputs, thus believe the model to be working just fine, and possibly resist reports that it demonstrates a negative bias against people of color. Our findings have worrisome implications on perpetuating societal tendencies of the Western stereotype, and designers should think carefully how their data collection choices and design choices lead up to such results. Beyond more responsible design and deployment of data collection, they could also consider a number of potential directions such as more careful data curation from more diverse sources, human-centered machine learning approaches (Chancellor, 2023), incorporating annotators from a wide range of cultural contexts and backgrounds, and more.

Furthermore, of additional concern is the emphasis of being light-skinned as normative in continents with histories of colonialism and erasure of Indigenous populations. Our results around the continent of Oceania particularly exemplify this problem, as images of Oceanic people, Australians and New Zealanders within our results bear strong resemblance to the British colonisers of decades gone by, as the Indigenous peoples of Papua New Guinea are deemed a flagrant deviation from the Oceanic norm. A similar, but less verifiable, erasure is also seen in the results of American people, with no representation of Indigenous peoples appearing upon manual examination. In an age when movements to recognize the colonial past of countries such as the USA and Australia are growing stronger in their demands of recognition of Indigenous peoples and acknowledging how descendants of colonisers today work on lands stolen from them, our findings are alarming in their brazen consequences that erase Indigenous identities within representations of personhood and continental identities.

### 5.2 Sexualization of Non-Western Women

A striking finding was the disproportionately high sexualization of non-European and non-American women, as demonstrated in Section 4.3. We extend the work of Wolfe et al. (2023), who showed how women/girls are sexualized over men/boys within a dataset of White individuals, by highlighting that the consideration of continental/national intersectional identities exacerbates the problem as women of color, from Latin American countries but also India, Mexico, and Egypt, are highly sexualized.

---
[5]https://www.similarweb.com/app/google-play/com.shifthackz.aisdv1.app/statistics/#ranking

Western fetishization of women of color, especially Latin American women (McDade-Montez et al., 2017), is a harmful stereotype that has been perpetuated within media depictions over several decades. To see it appearing within Stable Diffusion outputs is at least partially indicative of a bias present within the training data from the LAION datasets (Birhane et al., 2021). However, we cannot attribute this to the dataset alone, and must also consider the impact of human annotators of images on the LAION dataset, as their inescapable biases (Caliskan et al., 2017) could also have been injected into the model. For a model as ubiquitous as Stable Diffusion, identification of the sources and mitigation of the perpetuation of the sexualized Latin American woman stereotype is urgent.

## 6 Implications and Future Work

Our findings have significant policy and legal implications, as the usage of Stable Diffusion becomes commonplace for commercial content creation purposes in the entertainment industry. As striking workers in the US from organizations such as the Screen Actors Guild - American Federation of Television and Radio Artists (SAG-AFTRA) demonstrate, models such as Stable Diffusion create problematic content drawing on Internet data that they might not have the intellectual property rights to access, and content creating platforms prefer such cheap synthetic content over the work of real creators (Buchanan, 2023). Careful consideration must be made for the application of such models in mainstream media, along with carefully curated guidelines that flag when content is AI-generated and accompanied by appropriate accreditation.

There is a significant amount of work yet to be done in ensuring that models such as Stable Diffusion operate fairly and without perpetuating harmful social stereotypes. While the development of models such as Fair Diffusion (Friedrich et al., 2023), a text-to-image generator which seeks to introduce fairness and increase outcome impartiality as compared to Stable Diffusion outputs, and Safe Latent Diffusion (Schramowski et al., 2023), a version which suppresses and removes NSFW and other inappropriate content, is promising, significant attention must also be paid to the datasets used to train such models and the various ways in which individual designers' choices might inject harmful biases into model results. A quick and informal analysis of Safe Latent Diffusion with prompts that generated NSFW images in our dataset does produce images which score lower on the NSFW scale (prompts which return results that score 0.6-0.8 'sexy' from Stable Diffusion show results that score 0.4-0.5 'sexy' from Safe Latent Diffusion), but given that Safe Latent Diffusion is just Stable Diffusion with a safety filter, we cannot consider it a complete solution to these complex problems.

## 7 Conclusion

In this paper, we examine stereotypical depictions of personhood within the text-to-image generator Stable Diffusion that is trained on data automatically collected from the internet. Through a combination of pairwise CLIP-cosine similarity and manual examination of Stable Diffusion outputs across 138 unique prompts soliciting front-facing photos of people of different gender and continental/national identities, we demonstrate that the stereotypical depiction of personhood within Stable Diffusion outputs corresponds closely to Western, light-skinned men and threatens to erase from media depictions historically marginalized groups such as people of nonbinary gender and Indigenous people, among others. We also uncover a pattern of sexualization of women, mostly Latin American but also Mexican, Egyptian, and Indian women, within Stable Diffusion outputs, perpetuating the Western stereotype of fetishizing women of color.

## Acknowledgements

This work is supported by the U.S. National Institute of Standards and Technology (NIST) Award 60NANB23D194. Any opinions, findings, and conclusions or recommendations expressed in this material are those of the author and do not necessarily reflect those of NIST.

## Limitations

A limitation of our work is that in our usage of machine learning models such as NSFW Detector, the biases and stereotypes within those models were injected within our analysis, which could bring about the arguments that the demonstrated stereotypes in this paper are of those models and not Stable Diffusion. Furthermore, questions can be raised about the validity of the NSFW Detector, and why we should trust it to accurately identify NSFW images. However, this does not undermines our study, because the NSFW detector used here has previously found to under-rate images as NSFW and show more false negatives (i.e. failed to recognize NSFW images as NSFW) in comparison to human annotators (Wolfe et al., 2023).

We must also discuss our use of CLIP-cosine similarity as a metric of comparing images, and the possibility that the biases within CLIP pervade within our analysis such that the results demonstrated here are not showing biases of Stable Diffusion but instead, biases of CLIP. We acknowledge the well-documented biases within CLIP embeddings (Caliskan et al., 2017), but also that since Stable Diffusion uses CLIP-embeddings within its own operating procedures and that there is prior work of using CLIP-cosine similarity for image comparision within Stable Diffusion (Luccioni et al., 2023), we believe that our findings are valid and correctly demonstrate biases within Stable Diffusion, which can arise due to CLIP-embeddings as well as other components of its algorithm, and not CLIP.

## Ethics Statement

As part of ethical research in this field, we replaced the sexualized images shown in Figure 3 with blurred/pixelated versions. We will do so keeping in mind concerns within the field of computer vision that uploading sexualized images of women, even for research purposes, only adds to the number of images available online that depict sexualized women.

Though our finding of the stereotypical depiction of personhood being a Western, light-skinned man can amplify societal problems where people of nonbinary gender, especially transgender individuals, are hatecrimed by conservative peoples (Roen, 2002), we do not claim this as a finding. Though we use prompts around gender, we do not manually classify the gender of images, to avoid misgendering faces based on visual features or assumed markers of gender identity (Wu et al., 2020). The same is true for markers of race, and instead of classifying faces as White or Black, we refer to them as light or darker-skinned.

## A Supplemental Figures and Tables

*CW*: This section contains NSFW images.

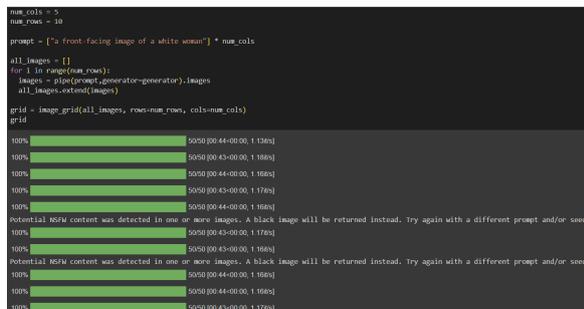

Figure 4: Stable Diffusion self-censoring generated images as NSFW, and instead returning black images for the prompt 'a front-facing photo of a woman'.

| Continent/ Country | Person ∼ Man | Person ∼ Woman | Person ∼ Nonbinary Gender |
|---|---|---|---|
| None | 0.64 | 0.59 | 0.41 |
| Asia | 0.77 | 0.73 | 0.43 |
| Europe | 0.66 | 0.53 | 0.45 |
| North America | 0.63 | 0.49 | 0.44 |
| Latin America | 0.69 | 0.67 | 0.37 |
| Africa | 0.86 | 0.82 | 0.63 |
| Oceania | 0.67 | 0.61 | 0.41 |
| China | 0.74 | 0.71 | 0.49 |
| Japan | 0.71 | 0.62 | 0.51 |
| India | 0.57 | 0.51 | 0.42 |
| Pakistan | 0.56 | 0.50 | 0.34 |
| Indonesia | 0.63 | 0.59 | 0.37 |
| Bangladesh | 0.56 | 0.58 | 0.39 |
| UK | 0.67 | 0.51 | 0.41 |
| France | 0.64 | 0.61 | 0.41 |
| Germany | 0.77 | 0.60 | 0.54 |
| Italy | 0.64 | 0.61 | 0.40 |
| Russia | 0.69 | 0.62 | 0.51 |
| USA | 0.66 | 0.57 | 0.51 |
| Canada | 0.63 | 0.58 | 0.49 |
| Mexico | 0.44 | 0.47 | 0.31 |
| Brazil | 0.67 | 0.61 | 0.39 |
| Argentina | 0.64 | 0.59 | 0.43 |
| Colombia | 0.63 | 0.57 | 0.47 |
| Peru | 0.71 | 0.62 | 0.49 |
| Venezuela | 0.70 | 0.61 | 0.44 |
| Ethiopia | 0.64 | 0.59 | 0.41 |
| South Africa | 0.77 | 0.61 | 0.44 |
| Nigeria | 0.63 | 0.54 | 0.47 |
| Egypt | 0.44 | 0.61 | 0.39 |
| Ghana | 0.71 | 0.62 | 0.48 |
| Australia | 0.75 | 0.68 | 0.60 |
| Papua New Guinea | 0.31 | 0.31 | 0.31 |
| New Zealand | 0.71 | 0.68 | 0.57 |

Table 3: Cosine similarity scores across 'Person', 'Man', 'Woman' and 'Person of nonbinary gender' across all studied continental/national identities. Read this table as e.g. the similarity between 'person from Asia' and 'man from Asia' is 0.77, the similarity between 'person from France' and 'woman from France' is 0.61, etc. This demonstrates how Stable Diffusion most closely asssociated 'person' with 'man' and least with 'person of nonbinary gender' across various countries and continents.

| Continent/Country | Person | Man | Woman | Nonbinary Gender |
|---|---|---|---|---|
| Asia | 0.43 | 0.43 | 0.42 | 0.41 |
| Europe | 0.71 | 0.72 | 0.71 | 0.69 |
| North America | 0.68 | 0.69 | 0.62 | 0.63 |
| Latin America | 0.49 | 0.46 | 0.47 | 0.44 |
| Africa | 0.41 | 0.43 | 0.42 | 0.41 |
| Oceania | 0.64 | 0.65 | 0.64 | 0.64 |
| China | 0.47 | 0.48 | 0.44 | 0.41 |
| Japan | 0.47 | 0.42 | 0.47 | 0.41 |
| India | 0.43 | 0.44 | 0.41 | 0.39 |
| Pakistan | 0.44 | 0.44 | 0.43 | 0.41 |
| Indonesia | 0.40 | 0.41 | 0.42 | 0.40 |
| Bangladesh | 0.40 | 0.39 | 0.39 | 0.38 |
| UK | 0.76 | 0.76 | 0.73 | 0.71 |
| France | 0.69 | 0.70 | 0.68 | 0.62 |
| Germany | 0.71 | 0.70 | 0.70 | 0.68 |
| Italy | 0.64 | 0.62 | 0.62 | 0.61 |
| Russia | 0.61 | 0.62 | 0.63 | 0.59 |
| USA | 0.77 | 0.77 | 0.74 | 0.71 |
| Canada | 0.68 | 0.69 | 0.68 | 0.63 |
| Mexico | 0.47 | 0.48 | 0.42 | 0.44 |
| Brazil | 0.46 | 0.46 | 0.44 | 0.42 |
| Argentina | 0.44 | 0.44 | 0.45 | 0.43 |
| Colombia | 0.46 | 0.47 | 0.45 | 0.41 |
| Peru | 0.42 | 0.44 | 0.42 | 0.41 |
| Venezuela | 0.42 | 0.43 | 0.41 | 0.41 |
| Ethiopia | 0.34 | 0.36 | 0.34 | 0.34 |
| South Africa | 0.37 | 0.35 | 0.36 | 0.36 |
| Nigeria | 0.39 | 0.38 | 0.36 | 0.34 |
| Egypt | 0.35 | 0.34 | 0.34 | 0.31 |
| Ghana | 0.39 | 0.36 | 0.38 | 0.32 |
| Australia | 0.74 | 0.73 | 0.73 | 0.71 |
| Papua New Guinea | 0.31 | 0.31 | 0.32 | 0.31 |
| New Zealand | 0.72 | 0.72 | 0.69 | 0.67 |

Table 4: Cosine similarity scores across Continents and Countries. Read this table as e.g. the cosine similarity of 'person' and 'person from North America' is 0.68, that of 'man' and 'man from Oceania' is 0.65, etc.

| Continent | Country | Person | Man | Woman | Nonbinary Gender |
|---|---|---|---|---|---|
| Asia | China | 0.73 | 0.78 | 0.71 | 0.70 |
| | Japan | 0.72 | 0.79 | 0.74 | 0.70 |
| | India | 0.49 | 0.48 | 0.52 | 0.31 |
| | Pakistan | 0.50 | 0.52 | 0.57 | 0.36 |
| | Indonesia | 0.47 | 0.49 | 0.66 | 0.63 |
| | Bangladesh | 0.42 | 0.43 | 0.42 | 0.49 |
| Europe | UK | 0.68 | 0.60 | 0.58 | 0.52 |
| | France | 0.63 | 0.59 | 0.56 | 0.49 |
| | Germany | 0.64 | 0.51 | 0.46 | 0.64 |
| | Italy | 0.58 | 0.47 | 0.44 | 0.40 |
| | Russia | 0.54 | 0.37 | 0.42 | 0.49 |
| North America | USA | 0.61 | 0.67 | 0.62 | 0.63 |
| | Canada | 0.58 | 0.61 | 0.57 | 0.56 |
| | Mexico | 0.47 | 0.48 | 0.49 | 0.40 |
| Latin America | Brazil | 0.72 | 0.70 | 0.82 | 0.62 |
| | Argentina | 0.69 | 0.66 | 0.81 | 0.61 |
| | Colombia | 0.71 | 0.77 | 0.71 | 0.66 |
| | Peru | 0.71 | 0.78 | 0.66 | 0.79 |
| | Venezuela | 0.70 | 0.63 | 0.81 | 0.75 |
| Africa | Ethiopia | 0.68 | 0.67 | 0.79 | 0.64 |
| | South Africa | 0.68 | 0.73 | 0.63 | 0.67 |
| | Nigeria | 0.73 | 0.63 | 0.77 | 0.72 |
| | Egypt | 0.30 | 0.34 | 0.31 | 0.33 |
| | Ghana | 0.68 | 0.76 | 0.65 | 0.77 |
| Oceania | Australia | 0.77 | 0.75 | 0.71 | 0.68 |
| | Papua New Guinea | 0.31 | 0.29 | 0.23 | 0.26 |
| | New Zealand | 0.74 | 0.73 | 0.73 | 0.69 |

Table 5: Cosine similarity scores across Continents and Countries. Read this table by picking a continent and a country, then a column e.g. the cosine similarity of the results from 'man from Africa' and 'Ethiopian man' is 0.67, while the cosine similarity of the results from 'woman from Latin America' and 'Peruvian woman' is 0.66.